\documentclass[10pt,twocolumn,letterpaper]{article}

\usepackage{cvpr}
\usepackage{times}
\usepackage{epsfig}
\usepackage{graphicx}
\usepackage{amsmath}
\usepackage{amssymb}

\usepackage{url}
\usepackage[ruled,vlined,linesnumbered]{algorithm2e}
\usepackage[T1]{fontenc}
\usepackage[utf8]{inputenc}
\usepackage[]{caption}
\usepackage{multirow}
\usepackage{pdfpages}
\usepackage{mathtools}% Loads amsmath
\usepackage{color}
\usepackage{soul}
\usepackage{booktabs}

\usepackage{pifont}% http://ctan.org/pkg/pifont
\newcommand{\xmark}{\text{\ding{55}}}

\usepackage[toc,page]{appendix}
\usepackage{chngcntr}

\usepackage[pagebackref=true,breaklinks=true,letterpaper=true,colorlinks,bookmarks=false]{hyperref}

\cvprfinalcopy % *** Uncomment this line for the final submission

\def\cvprPaperID{6223} % *** Enter the CVPR Paper ID here

\ifcvprfinal\fi
\begin{document}

%%%%%%%%% TITLE
\title{Distilling Effective Supervision from Severe Label Noise}

\author{Zizhao Zhang, Han Zhang, Sercan \"{O}. Ar{\i}k, Honglak Lee, Tomas Pfister \\
Google Cloud AI, Google Brain\\
%\texttt{\{zizhaoz,zhanghan,soarik,honglak,tpfister\}@google.com} 
}

\maketitle
%\thispagestyle{empty}

%%%%%%%%% ABSTRACT 

\begin{abstract}
Collecting large-scale data with clean labels for supervised training of neural networks is practically challenging. 
Although noisy labels are usually cheap to acquire, existing methods suffer a lot from label noise. This paper targets at the challenge of robust training at high label noise regimes. 
The key insight to achieve this goal is to wisely leverage a small trusted set to estimate exemplar weights and pseudo labels for noisy data in order to reuse them for supervised training. 
We present a holistic framework to train deep neural networks in a way that is highly invulnerable to label noise. 
Our method sets the new state of the art on various types of label noise and achieves excellent performance on large-scale datasets with real-world label noise.
For instance, on CIFAR100 with a 40\% uniform noise ratio and only 10 trusted labeled data per class, our method achieves $80.2{\pm}0.3\%$ classification accuracy, where the error rate is only 1.4\% higher than a neural network trained without label noise. 
Moreover, increasing the noise ratio to $80\%$, our method still maintains a high accuracy of $75.5{\pm}0.2\%$, compared to the previous best accuracy $48.2\%$\footnote{Source code available: \url{https://github.com/google-research/google-research/tree/master/ieg}}.
\end{abstract}

\section{Introduction}

Training deep neural networks usually requires large-scale labeled data. 
However, the process of data labeling by humans is challenging and expensive in practice, especially in domains where expert annotators are needed such as medical imaging. Noisy labels are much cheaper to acquire (e.g., by crowd-sourcing, web search, etc.). 
Thus, a great number of methods have been proposed to improve neural network training from datasets with noisy labels to take advantage of the cheap labeling practices \cite{zhang2018generalized}. 
However, deep neural networks have high capacity for memorization. When noisy labels become prominent, deep neural networks inevitably overfit noisy labeled data \cite{zhang2016understanding,tanaka2018joint}. 

\begin{figure}[t]
\begin{minipage}[b]{0.49\textwidth}
    \centering
      \includegraphics[width=0.99\linewidth]{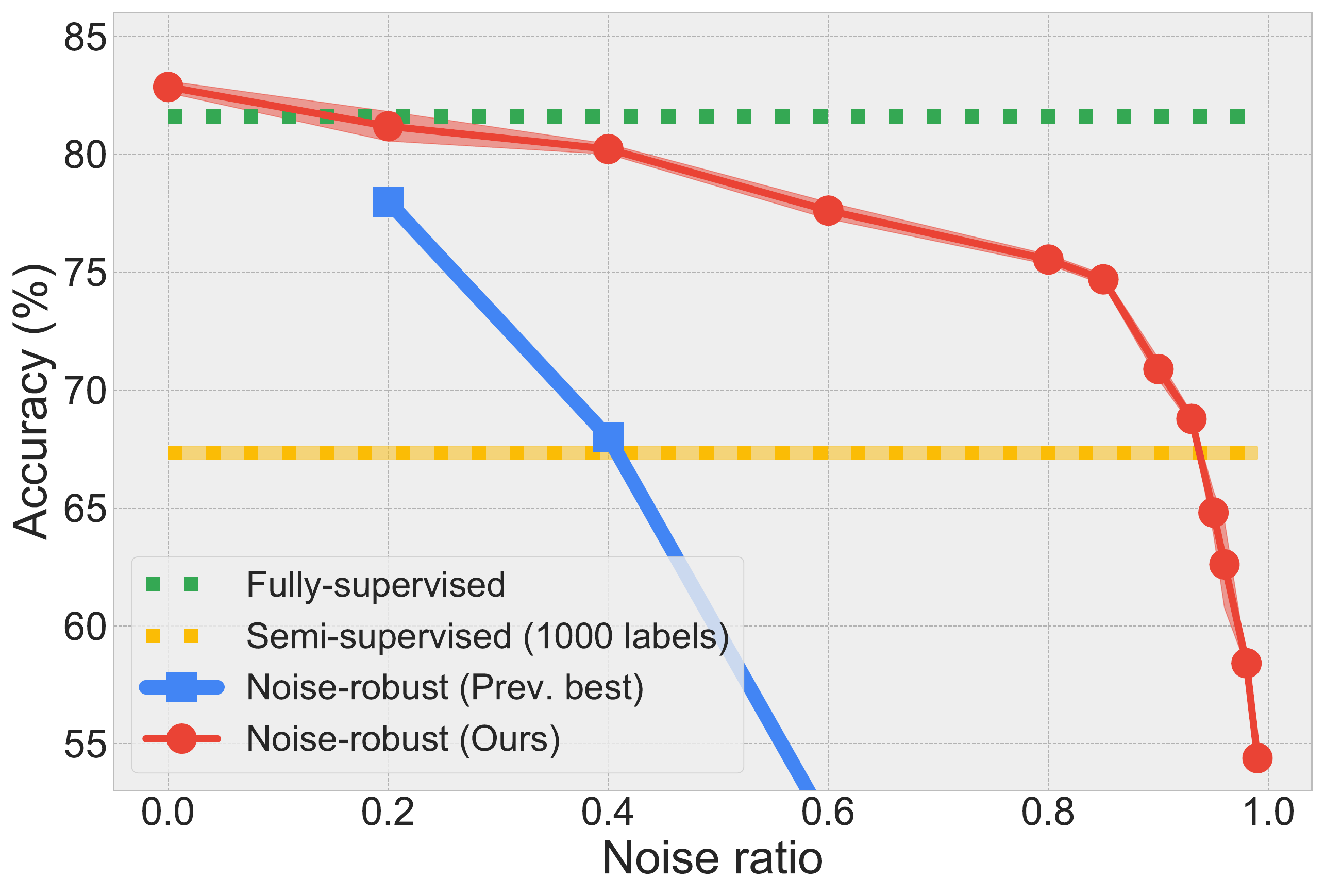}
\end{minipage}
\hfill
\begin{minipage}[b]{0.49\textwidth}
\centering
\begin{tabular}{c|ccccccc}
\toprule
 Ratio   & 0.85 & 0.9 & 0.93 & 0.95 & 0.96 & 0.98 & 0.99   \\ \toprule
  mean &   74.7 & 70.9 & 68.8 & 64.8 & 62.6 & 58.4 & 54.4 \\ \bottomrule
  %std &  0.21 & 0.45 & 0.26 & 0.91 & 1.85 & 0.16 & 0.29 \\ 
\end{tabular}
\end{minipage}
\caption{Image classification results on CIFAR100. 
  \textsf{Fully-supervised} denotes a model trained with all data without label noise.
  \textsf{Noise-robust (prev. best)} denotes the previous best results for noisy labels (50 trusted data per class are used by this method). 
  10 trusted data per class are available for \textsf{Semi-supervised} and \textsf{Noise-robust (ours)}. 
  The bottom table provides the accuracy of settings over 80\% noise ratios.
  \textsf{Semi-supervised} is our improved version of MixMatch~\cite{berthelot2019mixmatch}. Our method outperforms \textsf{Semi-supervised} at up to a 95\% noise ratio. The bottom table shows mean accuracy of three runs. See Section \ref{sec:compare_semi} for more details.
  }
  \vspace{-.2cm}
\label{fig:overview}
\end{figure}

To overcome this problem, we argue that building the dataset wisely is necessary. 
Most methods consider the setting where the entire training dataset is acquired with the same labeling quality. 
However, it is often practically feasible to construct a small dataset with human-verified labels, in addition to a large-scale noisy training dataset. If the methods based on this setting can demonstrate high robustness to noisy labels, new horizons can be opened in data labeling practices \cite{lee2018cleannet,xiao2015learning}.
There are a few recent methods that demonstrate good performance by leveraging a small trusted dataset while training on a large noisy dataset, including learning weights of training data \cite{jiang2017mentornet,ren2018learning}, loss correction \cite{hendrycks2018using}, and knowledge graph \cite{li2017learning}. However, these methods either require a substantially large trusted set or become ineffective at high noise regimes. 
In contrast, our method maintains superior performance with remarkably smaller size of the trusted set
(e.g., the previous best method \cite{jiang2017mentornet} uses up to 10\% of the total training data while our method achieves superior results with as low as 0.2\%). 

Given a small trusted dataset and large noisy dataset, there are two common machine learning approaches to train neural networks. The first is noise-robust training, which needs to handle label noise effects as well as distill correct supervision from the large noisy dataset. Considering the possible harmful effects from label noise, the second approach is semi-supervised learning, which discards noisy labels and treats the noisy dataset as a large-scale unlabeled dataset. 
In Figure \ref{fig:overview}, we compare methods of the two directions under such setting. We can observe that the advanced noise-robust method is inferior to semi-supervised methods even with a 50\% noise ratio (i.e., they cannot utilize the many correct labels from the other data), motivating the necessity for further investigation of noise-robust training. 
This also raises a practically interesting question: Should we discard noisy labels and opt in semi-supervised training at high noise regimes for model deployment? 

\textbf{Contributions:}
In response to this question, we propose a highly effective method for noise-robust training. 
Our method wisely takes advantage of a small trusted dataset to optimize exemplar weights and labels of mislabeled data in order to distill effective supervision from them for supervised training.
To this end, we generalize a meta re-weighting framework and propose a new meta re-labeling extension, which incorporates conventional pseudo labeling into meta optimization. 
We further utilize the probe data as anchors to reconstruct the entire noisy dataset using learned data weights and labels and thereby perform supervised training.
Comprehensive experiments show that even with extremely noisy labels, our method demonstrates greatly superior robustness compared to previous methods (Figure \ref{fig:overview}).
Furthermore, our method is designed to be model-agnostic and generalizable to a variety of label noise types as validated in experiments. 
Our method sets new state of the art on CIFAR10 and CIFAR100 by a significant margin and achieves excellent performance on the large-scale WebVision, Clothing1M, and Food101N datasets with real-world label noise.

\section{Related Work}
In supervised training, overcoming noisy labels is a long-term problem \cite{frenay2013classification,wang2019symmetric,li2019learning,ma2018dimensionality,yi2019probabilistic}, especially important in deep learning. Our method is related to the following discussed methods and directions.

Re-weighting training data has been shown to be effective \cite{liu2015classification}. However, estimating effective weights is challenging. \cite{ren2018learning} proposes a meta learning approach to directly optimize the weights in pursuit of best validation performance. \cite{jiang2017mentornet} alternatively uses teach-student curriculum learning to weigh data. \cite{han2018co} uses two neural networks to co-train and feed data to each other selectively. \cite{arazo2019unsupervised} models per sample loss and corrects the loss weights. 
Another direction is modeling confusion matrix for loss correction, which has been widely studied in \cite{sukhbaatar2014training,natarajan2013learning,tanno2019learning,patrini2017making,arazo2019unsupervised}. 
For example, \cite{hendrycks2018using} shows that using a set of trusted data to estimate the confusion matrix has significant gains. 

The approach of estimating pseudo labels of noisy samples is another direction and has a close relationship with semi-supervised learning \cite{li2017learning,tanaka2018joint,veit2017learning, han2019deep,lee2013pseudo,sohn2020fixmatch,pham2020meta}. 
Along this direction, \cite{reed2014training} uses bootstrapping to generate new labels. \cite{li2019learning} leverages the popular MAML meta framework \cite{finn2017model} to verify all label candidates before actual training.
Besides pseudo labels, building connections to semi-supervised learning has been recently studied \cite{kim2019nlnl}. 
For example, \cite{hataya2019unifying} proposes to use mixup to directly connect noisy and clean data, which demonstrates the importance of regularization for robust training. 
\cite{hataya2019unifying,arazo2019unsupervised} uses mixup \cite{zhang2017mixup} to augment data and demonstrates clear benefits. 
\cite{ding2018semi,kim2019nlnl} identifies mislabeled data first and then conducts semi-supervised training. 

%%Moreover, to potentially make use of information from mislabeled data. Previous methods have tried to online update labels using current models and bootstrap the model training with online estimated pseudo labels \cite{tanaka2018joint,reed2014training}. 

\section{Background}
%Correct data weights and labels are important for noise-robustness training.
Reducing the loss weight of mislabeled data has been shown effective in noise-robust training. 
Here we briefly introduce a meta learning based re-weighting (L2R) method \cite{ren2018learning}, serving as a base for the proposed method.
%\subsection{Meta Re-weighting}
L2R is a re-weighting framework that optimizes the data weights in order to minimize the loss of an unbiased trusted set matching the test data. The formulation can be briefly summarized as following.

Given a dataset of $N$ inputs with noisy labels $D_u = \{(x_i, y_i), 1 < i < N \}$ and also a small dataset of $M$ of samples with trusted labels $D_p = \{(x_i, y_i), 1 < i < M \}$ (i.e., probe data), where $M \ll N$. 
The objective function of training neural networks can be represented as a weighted cross-entropy loss:
\begin{equation}
\Theta^*(\mathbf{\omega}) = \arg \min_{\Theta} \sum_{i=1}^N \mathbf{\omega}_i L(y_i, \Phi(x_i; \Theta)),
\label{eq:weightobj}
\end{equation}
where $\mathbf{\omega}$ is a vector that its element $\mathbf{\omega}_i$ gives the weight for the loss of one training sample. $\Phi(\cdot; \Theta)$ is the targeting neural network (with parameters $\Theta$) that outputs the class probability and $L(y_i, \Phi(x_i; \Theta))$ is the standard softmax cross-entropy loss for each training data pair $(x_i, y_i)$.
We omit $\Theta$ in $\Phi(x_i; \Theta)$ frequently for conciseness.

The above is a standard weighted supervised training loss. L2R converts $\mathbf{\omega}$ as learnable parameters, and formulates a meta learning task to learn optimal $\mathbf{\omega}$ for each training data in $D_u$, such that the trained model using Equation \eqref{eq:weightobj} can minimize the error on a small and trusted dataset $D_p$ \cite{ren2018learning}, measured by the cross-entropy loss $L^p$ on $D_p$. 
The problem can be solved by repeatedly finding a combination of $\mathbf{\omega}$ that the trained model performs best.
However, it is computationally infeasible to compute since each update step of it requires training the model until converge before measuring $L^p$. 
In practice, it is possible to use an online approximation \cite{ren2018learning,finn2017model} to perform a single meta gradient-descent step $\Theta_{t+1}(\mathbf{\omega}) = \Theta_{t} - \alpha \nabla_{\Theta} \sum_{i}^{N} \mathbf{\omega}_i L\big(y_i, \Phi(x; \Theta_t)\big)$, where $\alpha$ is the step size. Therefore, the meta optimization of $\omega$ is defined as
\begin{equation}
\begin{split}
\mathbf{\omega}_t^* = \arg \min_{\mathbf{\omega}, \mathbf{\omega}\ge0} \frac{1}{M} \sum_{i}^{M} L^p\big(y_i, \Phi(x_i; \Theta_{t+1}(\mathbf{\omega}))\big), \\
s.t. \sum_j \mathbf{\omega}_{t,j} = 1.
\label{eq:weight}
\end{split}
\end{equation}
The re-weighting coefficients can be obtained by gradient descent $\mathbf{\omega}^* \approx \mathbf{\omega}_0 - \nabla_\mathbf{\omega} L^p |_{\mathbf{\omega} = \mathbf{\omega}_0}$ and then normalization to satisfy the constraints of $\mathbf{\omega}$ in Equation \eqref{eq:weight}.
The method expects that the optimized $\mathbf{\omega}^*$ coefficients should assign low weight values to mislabeled data to isolate mislabeled data from clean data. Note that since $\Theta_{t+1}(\mathbf{\omega})$ is a function of $\omega$, the optimization of $\mathbf{\omega}$ using $L^p$ requires second-order back-propagation (sometimes called gradient-by-gradient) \cite{ren2018learning}.

\section{Proposed Method}
Besides estimating exemplar weights from the noisy data, it is also important to estimate the correct labels via re-labeling process. We informally call this process as estimation of ``Data Coefficients” (i.e., exemplar weights and true labels), which are two major information for constructing supervised training. 
We present a generalized framework to estimate data coefficients via meta optimization. 

The motivation of studying re-labeling is straightforward. When the noise ratio is high, a significant amount of data would be discarded and thereby would make no contribution to the model training. 
To address this inefficiency, it is necessary to enable the reuse of mislabeled data to improve performance at high noise regimes. Different from pseudo labeling in semi-supervised learning \cite{lee2013pseudo}, a portion of labels in noisy datasets are correct. Thus, distilling them effectively bring extra benefits.
In contrast to previous pseudo labeling noise-robust methods \cite{li2019learning}, our proposed method constructs a differentiable pseudo re-labeling objective to select the best choice efficiently.

\subsection{Initial pseudo label estimator}
\label{sec:initiallabel}
Utilizing the pseudo labels for unlabeled training data is widely studied for semi-supervised learning \cite{lee2013pseudo,tanaka2018joint,lee2013pseudo}. Pseudo labels are usually inferred by the model predictions.
Neural networks can be unstable to input augmentations \cite{zheng2016improving,azulay2018deep}.
To generate more robust label guessing, a recent semi-supervised learning method \cite{berthelot2019mixmatch} considers averaging predictions over $K$ augmentations. 
We adopt this simple technique to initialize soft pseudo labels, which is given by averaging predictions of different input augmentations:
\begin{equation}
\begin{split}
g(x, \Phi)_i & =  Pr_i^\frac{1}{\tau}/ \sum_{j} Pr_j^\frac{1}{\tau}, \\
& \text{where } Pr = \frac{1}{K} \big(\Phi(x) + \sum_{k=1}^{K-1} \Phi(\hat{x}_k)\big)
\label{eq:pseudo-label}
\end{split}
\end{equation}
where $\hat{x}_k$ is $k$-th random augmentations of input $x$. $g(x)$ is the estimated pseudo label of $x$, where $g_i$ represents the $i$-th class probability. $\tau$ is a softmax temperature scaling factor used to sharpen the pseudo label distribution ($\tau = 0.5$ in this paper).

\subsection{Improved pseudo label initialization}
\label{sec:consistlabel}
To make pseudo labels effective for supervised training eventually, the distribution of pseudo labels needs to be sharp and consistent across augmented versions of inputs.
If the predictions of input augmentations are inconsistent to each other, averaging them with Equation \eqref{eq:pseudo-label} would cause their contributions to cancel out, yielding a flattened pseudo label distribution. 
From this insight, reducing the inconsistency of predictions of augmentations is necessary. Therefore, we propose to improve pseudo label estimation by incorporating a KL-divergence loss 
\begin{equation}
  \min_{\Theta} \; L_{\text{KL}} = \frac{1}{N} \sum_i^{N} \text{KL}\big(\Phi(x_i;\Theta) \,\big|\big|\, \Phi(\hat{x}_i;\Theta) \big),
\label{eq:consistlabel}
\end{equation}
which penalizes inconsistency of arbitrary input augmentations $\hat{x}_i$ of $x_i$. The effectiveness of this loss is studied in experiments.

\subsection{Meta re-labeling}
\label{sec:metarelabeling}
For each training data $x$, we now have initial pseudo label $g(x, \Phi)$ and its original label $y$. We formulate the problem of re-labeling as finding the best selection of the two candidates for each data efficiently to reduce the error of the probe data most. 
Based on the meta re-weighting idea~\cite{ren2018learning}, we propose a new objective that combines the estimation of data coefficients efficiently: %``data coefficient'' (i.e., exemplar weights and corrected labels) efficiently:
\begin{equation}
\begin{split}
&\Theta^* (\mathbf{\omega}, \lambda) = \arg \min_{\Theta} 
\sum_{i=1}^N \mathbf{\omega}_{i} L\big(\mathcal{P}(\lambda_{i}), \, \Phi(x_i; \Theta)\big), \\
&\mathcal{P}(\lambda_{i}) = \lambda_{i} y_i + (1-\lambda_{i}) g(x_i, \Phi) \;\; s.t. \; 0 \le \lambda_{i} \le 1,
\end{split}
\label{eq:theta}
\end{equation}
where $\mathcal{P}$ is a function of parameter $\lambda_{i}$ that is differentiable. In the meta step, $\lambda_{i}$ is designed to aggregate the original labels and the pseudo labels, which simplifies the back-propagation. 

Similar to how re-weighting works with second-order back-propagation, we can back-propagate the model using the loss $L^p$ on the probe data to optimize re-labeling coefficients $\lambda_{i}^*$.
In our implementation, we calculate the sign of its gradient for each data $x_i$ and rectify it:
%Different from~\cite{ren2018learning}, we calculate the sign of its gradient for each data $x_i$ and rectify it:
\begin{equation}
\begin{split}
%\lambda_{i}^* = \text{sign}\Big( - \frac{\partial } {\partial \lambda_{i}} \mathbb{E} \big[L^p|_{\lambda = \lambda_0, \mathbf{\omega} = \mathbf{\omega}_0} \big] \Big).
\lambda_{i}^* = \left[ \text{sign}\Big( - \frac{\partial } {\partial \lambda_{i}} \mathbb{E} \big[L^p|_{\lambda = \lambda_0, \mathbf{\omega} = \mathbf{\omega}_0} \big] \Big) \right]_+.
\end{split}
\label{eq:lambda}
\end{equation}
The motivation to use the (rectified) sign of the gradient instead of $\lambda \approx \lambda_0 - \nabla_\lambda L^p|_{\lambda = \lambda_0}$ (as how $\omega^*$ is calculated) are two folds: 1) $\nabla_\lambda L^p$ would become very small at later learning stage when pseudo labels are close to real labels (see Appendix A for mathematical illustration) and 2) simply aggregating $y_i$ and $g(x_i, \Phi)$ using scalar ($\lambda_0 - \nabla_\lambda L^p$) would make resulting pseudo label distribution not sufficiently sharp for supervised training.
Therefore, our method proposes to obtain the final pseudo labels as
\begin{equation}
y^*_i= 
\begin{cases}
    y_i,           & \text{if } \lambda_{i}^* > 0 \\
    g(x_i, \Phi),  & \text{otherwise}
\end{cases}
\end{equation}
After the meta step, we add two cross-entropy losses with respective to optimal $\mathbf{\omega}^*_i$ and $y_{i}^*$,
\begin{equation}
\vspace{-.1cm}
\begin{split}
L_{\mathbf{\omega}^*} & =  \sum_i^N \mathbf{\omega}^*_{i} L\big( \mathcal{P}(\lambda_{0}), \; \Phi(x_i; \Theta)\big), \\
L_{\lambda^*} & = \sum_i^N \mathbf{\omega}_{0} L\big(y^*_i, \; \Phi(x_i; \Theta)\big),
\end{split}
\vspace{-.1cm}
\end{equation}
%(\lambda_{t,i}^* < 0)[y,g(\Phi(x_i))]
%At the meta optimization step, L2R sets $\mathbf{\omega}_0=0$ to approximate the influence function \cite{koh2017understanding}, which results in $\nabla_{\Theta_{t}} \sum_{i}^{N} \mathbf{\omega}_i L\big(y_i, \Phi(x; \Theta_t)\big) \equiv 0$.
Similar to L2R, we use momentum SGD for model training. L2R sets $\mathbf{\omega}_0 = 0$ and uses naive gradient descent to estimate perturbation around $\mathbf{\omega}$. In contrast, we compute the meta step model parameters $\Theta_{t+1}$ by calculating the exact momentum update direction using momentum states of the SGD optimizer\footnote{For each training batch, we set initial the values as $\mathbf{\omega}_0 = 1/B$ (where $B$ is the batch size), treating each data equally. We use $\lambda_0=0.9$ (lean to original labels) based on the observation of better performance.}.

% global change
\SetKwInput{KwInput}{Input}
\SetKwInput{KwOutput}{Output}
\IncMargin{0.0em}
\begin{algorithm}[t]
% \SetAlgoLined
\KwInput{Current model parameters $\Theta^{t}$, A batch of training data $X_u$ from $D_u$, a batch of probe data $X_p$ from $D_p$, loss weight $k$ and $p$, threshold $T$ }
\KwOutput{Updated model parameters $\Theta^{t+1}$ }
Generate the augmentation $\hat{X}_u$ of $X_u$.\\
Estimate the pseudo labels via $g(x_u, \Phi), x_u \sim X_u{\cup}\hat{X}_u$ (Section \ref{sec:initiallabel} \& \ref{sec:consistlabel}).\\
Compute optimal data coefficients $\lambda^*$ and $\mathbf{\omega}^*$ via the meta step (Section \ref{sec:metarelabeling}).\\
Split the training batch $X_u$ (also corresponding $\hat{X}_u$) to possible clean batch $X_u^c$ and possible mislabeled batch $X_u^u$ using the binary criterion $\mathbb{I}(\mathbf{\omega}^* < T)$. \\
Construct the joint batch set (Section \ref{sec:sup}), 
$$X_p \cup X_u^u \cup X_u^c \cup \hat{X}_u^u \cup \hat{X}_u^c,$$
where $\hat{X}_u^u \cup X_u^u$ uses pseudo labels estimated by $g(\cdot, \Phi)$. \\
Compute the total loss for model update
$$L_{\mathbf{\omega}^{*}} + L_{\lambda^{*}} + L_{\beta}^p + p \ L_{\beta}^u + k \ L_{\text{KL}}.$$\\
Conduct one step stochastic gradient descent to obtain $\Theta^{t+1}$. \\
\caption{A training step of our method at time step $t$} \label{alg}
\end{algorithm}

\begin{table*}[t]
\caption{Validation accuracy on CIFAR10 with uniform noise. $M$ denotes the number of trusted (probe) data used. 0.01k indicates 1 image per class. For reference, vanilla training of WRN28-10/ResNet29 leads to 96.1\%/92.7\% accuracy. $^*$ indicates results trained by us.} \label{tab:cifar10uniform}
\vspace{-.2cm}
\centering
\begin{tabular}{lc|cccc}
 \toprule
 \multirow{2}{*}{Method}   & \multirow{2}{*}{$M$}  & \multicolumn{4}{c}{Noise ratio} \\ \cmidrule{3-6}
                           &                    & 0 & 0.2 & 0.4 & 0.8 \\ \midrule
 GCE \cite{zhang2018generalized}      & -  & 93.5 & 89.9$\pm$0.2 & 87.1$\pm$0.2 & 67.9$\pm$0.6 \\
 MentorNet DD \cite{jiang2017mentornet}  & 5k  & 96.0 & 92.0 & 89.0 & 49.0   \\
 RoG \cite{lee2019robust}       & -  &   94.2   & 87.4  & 81.8 &  -  \\ 
 L2R \cite{ren2018learning}     & 1k & 96.1 & 90.0$\pm0.4^*$ & 86.9$\pm$0.2 & 73.0$\pm0.8^*$ \\ 
 Arazo \textit{et al.} \cite{arazo2019unsupervised}    &  -   & 93.6  & 94.0   &   92.0  & 86.8 \\ 
 \midrule
%  Luo \cite{}       & - & 96.2 & 96.1 & 95.0$\pm$0.2 & 86.7$\pm$0.5 \\ \hline
 
 Ours-RN29 & 0.1k &  94.4 & 92.9$\pm$0.2 & 92.5$\pm$0.5 & 85.6+1.1   \\ 
 Ours    & 0.01k & 96.8 & 95.4$\pm$0.6 & 94.5$\pm$1.0 & 87.9$\pm$5.1 \\
 Ours    & 0.05k & 96.8 & 96.4$\pm$0.0 & 95.5$\pm$0.6 & 91.8$\pm$3.0 \\
 Ours    & 0.1k & 96.8 & \textbf{96.2}$\pm$\textbf{0.2} & \textbf{95.9}$\pm$\textbf{0.2} & \textbf{93.7}$\pm$\textbf{0.5} \\ \bottomrule
\end{tabular}
\vspace{-.2cm}
\end{table*}

\begin{table*}[t]
\caption{Validation accuracy on CIFAR100 with uniform noise. Standard training of WRN28-10/RN29 leads to 81.6\%/71.3\% accuracy. 0.1k indicates 1 image per class.  $^*$ indicates results trained by us.} \label{tab:cifar100uniform}
\vspace{-.2cm}
\centering
\begin{tabular}{lc|cccc}
 \toprule
 \multirow{2}{*}{Method}   & \multirow{2}{*}{$M$} &\multicolumn{4}{c}{Noise ratio} \\ \cmidrule{3-6}
            &   & 0 & 0.2 & 0.4 & 0.8 \\ \midrule
 GCE \cite{zhang2018generalized}      & -  & 81.4 & 66.8$\pm$0.4 & 61.8$\pm$0.2 & 47.7$\pm$0.7 \\
 MentorNet \cite{jiang2017mentornet}  & 5k  & 79.0 & 73.0 & 68.0 & 35.0  \\
%  RoG \cite{lee2019robust}       & -  &   94.18   & 58.2  & - &  -  \\ 
 L2R \cite{ren2018learning}       & 1k & 81.2 & 67.1$\pm0.1^*$ & 61.3+2.0 & 35.1$\pm1.2^*$ \\
 Arazo \textit{et al.} \cite{arazo2019unsupervised}    & - & 70.3   &  68.7 &    61.7  & 48.2 \\  \midrule
 
%  Luo \cite{}       & - & 81.4 & 80.6$\pm$0.2 & 74.27$\pm$0.5 & 66.32$\pm$0.8 \\ \hline
 Ours-RN29   &  1k & 72.1 & 69.3$\pm$0.5 & 67.0$\pm$0.8 & 60.7$\pm$1.0 \\ 
 Ours       & 0.1k &  83.0 & 77.4$\pm$0.4 & 75.1$\pm$1.1 & 62.1$\pm$1.2 \\
 Ours       & 0.5k &  83.0 & 80.4$\pm$0.5 & 79.6$\pm$0.3 & 73.6$\pm$1.5 \\
 Ours       
 &  1k & 83.0 & \textbf{81.2}$\pm$\textbf{0.7} & \textbf{80.2}$\pm$\textbf{0.3} & \textbf{75.5}$\pm$\textbf{0.2}  \\ \bottomrule
\end{tabular}
\vspace{-.2cm}
\end{table*}

\begin{table}[t!]
    \caption{Asymmetric noise on CIFAR10. } \label{tab:cifar10asy}
    \vspace{-.2cm}
    \centering
    \begin{tabular}{l|ccc}
     \toprule
    \multirow{2}{*}{Method} & \multicolumn{3}{c}{Noise ratio} \\ \cmidrule{2-4}
               & 0.2 & 0.4 & 0.8 \\ \hline
    %  WRN28-10   & -  & 90.3+0.4 & 78.3$\pm$0.3 & ? \\
     GCE \cite{zhang2018generalized}        & 89.5$\pm$0.3 & 82.3$\pm$0.7 & - \\
     LC \cite{patrini2017making}    & 89.1$\pm$0.5 & 83.6$\pm$0.3 & -\\\midrule
    %  Yucen \cite{}   & 94.0$\pm$0.2 & 85.6$\pm$0.8 & - \\ \hline
     Ours-RN29  & 92.7$\pm$0.2 & 90.2$\pm$0.5 & 78.9$\pm$3.5 \\
     Ours        & \textbf{96.5}$\pm$\textbf{0.2} & \textbf{94.9}$\pm$\textbf{0.1} & \textbf{79.3}$\pm$\textbf{2.4}\\\bottomrule
    \end{tabular}
    \vspace{-.3cm}
\end{table}

\subsection{Supervised training}
\label{sec:sup}
Given estimated data coefficients using probe data, we further leverage the effectiveness of it to construct supervised training. 
When introducing probe data for supervised training, appropriate regularizations are important to prevent overfitting on the probe data and the consequent failure of meta optimization (i.e., when $L^p$ in Equation \eqref{eq:lambda} gets very small).

We divide the data as either possibly-mislabeled (which are assigned with pseudo labels) or possibly-clean (which are assigned with original labels) using the binary criterion $\mathbb{I}(\mathbf{\omega}_i < T)$, where $T$ is a scalar threshold.
We treat the probe data as anchors to pair each training data and apply mixup \cite{zhang2017mixup}.
In this way, the model never sees the original probe data directly but the interpolated point between probe and training data, which can reduce overfitting on the probe data.
In detail, we construct supervised cross-entropy losses on the mixed data in the form of convex combinations using the data and their labels given a mixup factor $\beta$: $\text{Mix}_{\beta}(a, b) = \beta a + (1-\beta)b,\; \beta \sim
 \text{Beta}(0.5, 0.5)$. 
In detail, for each data $x_a$ in the concatenated data pool in $D_p\cup \hat{D}_u\cup D_u$, we apply pairwise mixup between the input batch and its random permutation,
\begin{equation}
\begin{split}
   x_\beta & = \text{Mix}_{\beta}(x_a, x_b), \;\; y_\beta = \text{Mix}_{\beta}(y_a, y_b), \\ 
   & \text{where } \{(x_a, y_a), (x_b, y_b)  \in D_p\cup \hat{D}_u\cup D_u\},
\end{split}
\end{equation}
where $\hat{D}_u$ is the augmented copy of $D_u$ (which is used by Equation \eqref{eq:pseudo-label}).
In detail, we introduce two softmax cross-entropy losses: $L_{\beta}^{p}$ for resulting mixed data when $x_a \sim D_p$ is from probe data
and $L_{\beta}^{u}$ when $x_a \sim \hat{D}_u\cup D_u$. 
The experiments show that our approach can reduce the probe data size to one sample per class.

\subsection{End-to-end training process}
Our training approach is end-to-end in one stage. A single gradient descent step can be structured in three sub-steps, meta-optimize data coefficients, construct augmented data, and update the model using aggregated losses.
Algorithm \ref{alg} illustrates a complete training step and specifies the joint objectives and their coefficients. Appendix B discusses the training efficiency.

\section{Experiments}
\subsection{Implementation details and experimental setup}
%\subsection{Empirical training details}
Here we discuss training details and hyperparameters that are shown to be useful for our experiments. More training details can be found in the Appendix.

\begin{table}[t!]
        \caption{Experiments with semantic noise where labels are generated by a neural network trained on limited data. The resulting noise ratio is shown in parentheses.} \label{tab:cifarsematic}
        \vspace{-.2cm}
        \centering
        \begin{tabular}{l|c|cc}
        \toprule
        Method & CIFAR10 (34\%) & CIFAR100 (37\%)\\ \midrule
        RoG \cite{lee2019robust}  & 70.0  & 53.6  \\ 
        L2R$^{*}$ \cite{ren2018learning}  &  71.0     & 56.9  \\ \midrule
        Ours-RN29     &   81.8  & 65.1 \\
        Ours  & \textbf{88.3}    &   \textbf{73.7 }  \\\bottomrule
        \end{tabular}
        \vspace{-.3cm}

\end{table} 

\begin{figure}[t!]
     \centering
     \includegraphics[width=.8\linewidth]{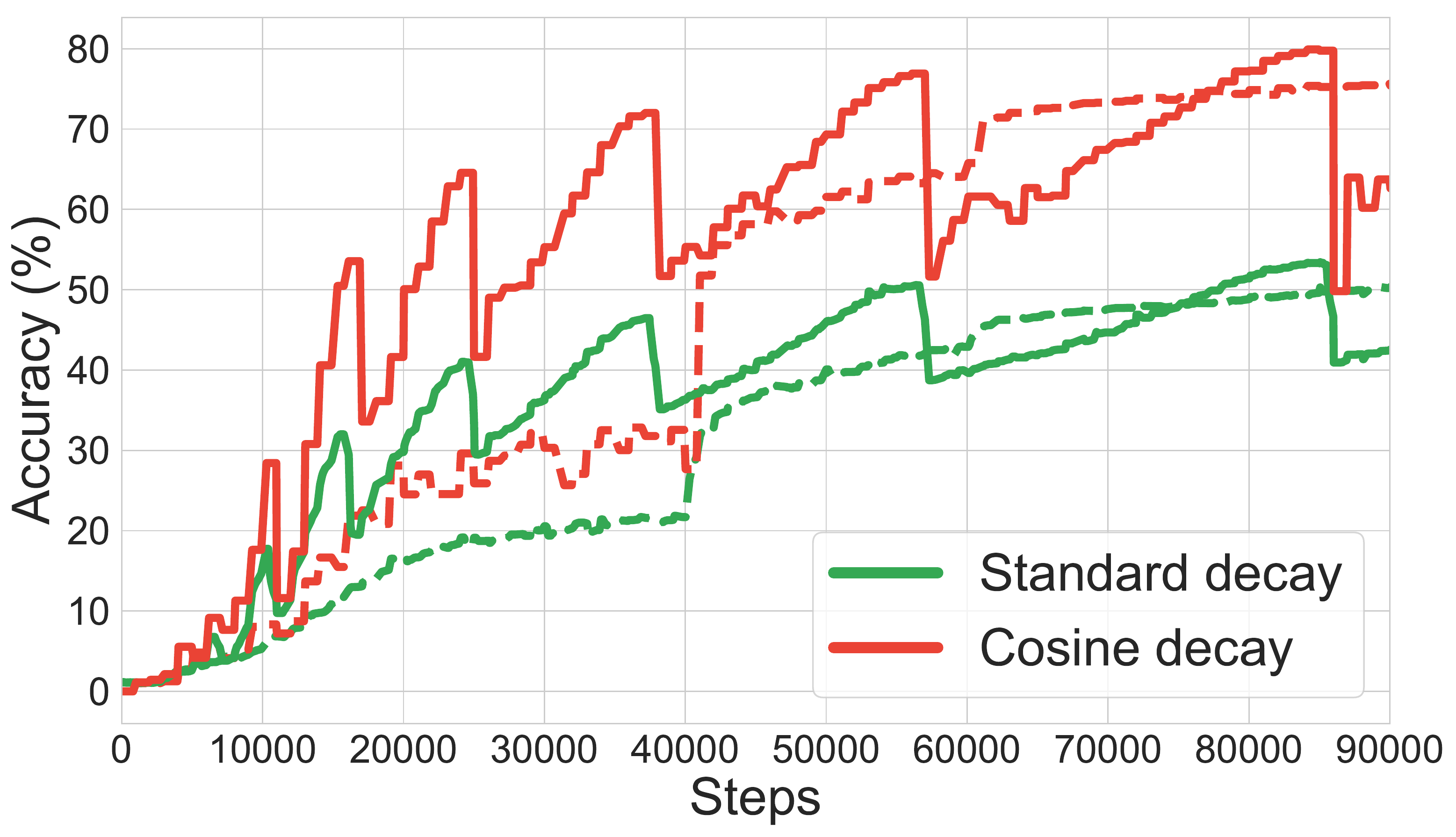}
     \vspace{-.2cm}
     \caption{Comparison with standard learning rate decay strategy. We use the commonly accepted setting (also used by L2R): the initial learning rate is 0.1, the learning rate decays to previous 0.1x at 40K and 50K steps. We show the training curves on CIFAR10 with 40\% uniform label noise. Dotted and solid lines are
     evaluation and training accuracy curves, respectively.}
     \label{fig:custom_step}
     \vspace*{-0.15in}
\end{figure}

\textbf{Model training:}
We adopt the Cosine learning rate decay with warm restarting \cite{loshchilov2016sgdr}\footnote{This learning rate schedule restarts from a larger value after each ``cosine'' cycle, so it yields a training curve with repeated `jag' shapes (see Figure \ref{fig:custom_step}). We set the initial cycle length to be one epoch, and after then cycle length increases by a factor of 1.5 and meanwhile the restart learning rate decreases by a factor of 0.9 as described in \cite{loshchilov2016sgdr}.}. 
In detail, we selected models at the lowest learning rate before the end of scheduled epochs for reporting result. 
We observe 3\%-5\% accuracy improvement on CIFAR datasets compared with the standard learning rate decay schedule (i.e., as used by L2R \cite{ren2018learning}), especially at large noise ratios. 
Figure~\ref{fig:custom_step} compares the training curves. 
Although it works particularly well in our method, we do not observe strong benefit for either training vanilla neural networks or training L2R. Further investigation are left as future work.

\begin{table*}[h]
    \caption{Open-set noise on CIFAR10. Each column indicates where the noisy out-of-distribution images are from. 
    RoG uses DenseNet-100 and L2R uses WRN28-10. 
    We run the baseline for better comparison (the first block of the table).} \label{tab:openset}
    \vspace{-.1cm}
    \centering
   \begin{tabular}{l|ccc}
    \toprule
     Method                      &   CIFAR100       &   CIFAR100+ImageNet & ImageNet      \\ \midrule
     RN29                      &      77.8  &	80.3 &	84.4           \\                      
     DenseNet-100 \cite{lee2019robust}                       &     79.0             &       86.7              &   81.6   \\
     WRN28-10                        &        82.8  &	84.7 &	88.7     \\   \midrule
     L2R  \cite{ren2018learning}                       &     81.8 &	81.3 &	85.0      \\    
     RoG  \cite{lee2019robust}                       &      83.4        &    87.1       &      84.4 \\  \midrule
     Ours-RN29                             &       86.4           &      87.4              &   90.0  \\  
     Ours                        &           \textbf{92.3}	& \textbf{93.0}	& \textbf{94.0}   \\       \bottomrule
\end{tabular}
\vspace{-.4cm}
\end{table*} 

\begin{table}[t]
 \caption{Large-scale WebVision experiments on mini and full versions. The top-1/top-5 accuracy on the ImageNet validation set are reported.} \label{tab:web} \vspace{-.2cm}
  \centering
   \begin{tabular}{c|cc}
    \toprule
     Method                                 & mini      & full \\ \bottomrule
     Co-teaching \cite{han2018co}             &  61.5/84.7 & -              \\
     Chen \textit{el al.} \cite{chen2019understanding}    & 61.6/85.0 & - \\
     MentorNet \cite{jiang2017mentornet}    & 63.8/85.8 & 64.2/84.8 \\  \bottomrule
     Ours-RN50                              & 78.0/94.4 & 65.8/85.8 \\
     Ours                                   & \textbf{80.0/94.9} & \textbf{69.0/88.3} \\     \bottomrule
\end{tabular}
\vspace{-.3cm}
\end{table}

\begin{table}[t]
    \caption{Food101N experiments. } \label{tab:food101n} \vspace{-.2cm}
  \centering
   \begin{tabular}{c|cc}
    \toprule
     Method                                 & Accuracy    \\ \bottomrule
     ResNet50 \cite{lee2017cleannet}    & 81.44     \\
     CleanNet \cite{lee2017cleannet}   &  83.95         \\
     Self-Learning \cite{han2019deep}   &  85.11         \\\bottomrule
     Ours-RN50                       &  \textbf{87.57} \\     \bottomrule
\end{tabular}
\vspace{-.3cm}
\end{table}

\textbf{Augmentation:}
Augmentation generates pixel-level perturbations on the original training inputs, which plays a critical role in Equation \eqref{eq:pseudo-label} and \eqref{eq:consistlabel}. 
We use the recently-proposed data augmentation technique based on policy-based augmentation (PA), AutoAugment \cite{cubuk2018autoaugment}, in our experiments. PA includes data processes of (policy augmentation${\rightarrow}$flip${\rightarrow}$random crop${\rightarrow}$cutout \cite{devries2017improved}). In detail, for each input image, we first generate one standard augmentation (random crop and horizontal flip) and then apply PA to generate $K$ random augmentations on top of the standard one. We fix $K=2$ augmentations in our experiments.
We further analyze the effects of learned policies and random policies (i.e., with no learning required) in Section~\ref{sec: discussion}.

\begin{table*}[t]
\caption{A comparison to semi-supervised methods and our semi-supervised extension (MixMatch-KL). MixMatch and MixMatch-KL$^*$ use WRN-28-2. 10 labeled data per class are used for semi-supervised training and the probe data of our method. Previous best scores for this task are compared.} \label{tab:cifar_semi_experiment}
\vspace{-.2cm}
\centering
\begin{tabular}{l|ccc|ccc}
 \toprule
    &       \multicolumn{3}{c|}{Semi-supervised}    & \multicolumn{2}{c}{Noise-robust (80$\%$ noise)} \\ \midrule
 Dataset                & MixMatch \cite{berthelot2019mixmatch} & MixMatch-KL$^*$ &  MixMatch-KL      & Prev. best \cite{arazo2019unsupervised}  &   Ours \\ \midrule
 CIFAR10            & 51.2  & 92.4$\pm$0.7  & 94.5$\pm$0.3 &   86.8    &   93.7$\pm$0.5 \\
 CIFAR100           & 34.5  & 57.6$\pm$0.4  & 67.3$\pm$0.3 &  48.2    &    75.2$\pm$0.2 \\
  \bottomrule
\end{tabular}
\vspace{-.3cm}
\end{table*}

\subsection{CIFAR noisy label experiments}
We follow \cite{ren2018learning,jiang2017mentornet} to conduct CIFAR10 and CIFAR100 experiments. 
For all CIFAR experiments with different noise types and ratios, we set $T=1, p=5, k=20$, which are empirically determined on CIFAR10 with 40\% uniform noise. Standard deviation are obtained over 3 runs with random seeds (and random data splits). We compare the proposed method against several recent methods, which have achieved leading performance on public benchmarks. Similar to L2R, we use the Wide ResNet (WRN28-10) \cite{zagoruyko2016wide} as default, unless specified otherwise for fair comparison. We also test our method using ResNet29 (RN29)\footnote{We follow this v2 implementation \url{https://github.com/keras-team/keras/blob/master/examples/cifar10_resnet.py}, 
which contains 0.84M parameters.}, 
which is much smaller than the ones used by compared methods. 

\textbf{Common random label noise}: 
Table \ref{tab:cifar10uniform} compares the results for CIFAR10 with uniform noise ratios of 0.2, 0.4, and 0.8. 
Our method yields 96.5\% accuracy at 20\% noise ratio and 94.7\% accuracy at 80\% noise ratio, demonstrating nearly noise-invulnerable performance. 
It still achieves the best performance with ResNet29.
We also train our full method with 0\% noise as reference. 
%These results show that our proposed method possibly provides additional forms of regularization to improve generalization.
Table \ref{tab:cifar100uniform} compares the results in CIFAR100 with uniform noise ratios of 0.2, 0.4, and 0.8. 
Additionally, we test our method with 10 images, 5 images and the extreme case of 1 image per class as probe data. 
MentorNet uses 5k clean images (50 per class) while our method 
reduces this number by up to 50x and maintains outperformed accuracy.

\textbf{Semantic label noise}: 
Next, we test our method on more realistic noisy settings on CIFAR.
By default, 10 images per class are used as probe data. 
First, Table \ref{tab:cifar10asy} compares the results on CIFAR10 with asymmetric noise ratios of 0.2, 0.4, and 0.8. 
Asymmetric noise is known as a more realistic setting because it corrupts semantically-similar classes (e.g., truck and automobile, or bird and airplane) \cite{patrini2017making}. 
Second, we follow RoG \cite{lee2019robust} to generate semantically noisy labels by using a trained VGG-13 \cite{simonyan2014very} on 5\% of CIFAR10 and 20\% of CIFAR100\footnote{We directly use the data provided by RoG authors. VGG-13 the hardest setting.}.
Table \ref{tab:cifarsematic} reports the compared results. 

\textbf{Synthetic open-set noise}:
Open-set is a unique type of noise that occurs in images rather than labels \cite{bendale2016towards,wang2018iterative}. We test our method on three kinds of synthetic open-set noisy labels provided by \cite{lee2019robust} in Table \ref{tab:openset}.
In all semantic noise settings, our method consistently outperforms the compared methods with a significant margin.  
From baseline comparison of supervised training in Table \ref{tab:openset}, we can see model capacity is beneficial for performance.
However, L2R, which uses WRN28-10, does not outperform its supervised WRN28-10, which implies that data re-weighting might not sufficient to deal with this noise type.
%Our method remains the best scores, demonstrating constant robustness.  
%The results indicate that our method is label corruption type agnostic.

\subsection{Large-scale real-world experiments}
WebVision \cite{li2017webvision} is a large-scale dataset which consists of real-world noisy labels. 
%The images are obtained by crawling from the Flickr website and Google Images Search using ImageNet labels.
It contains 2.4 million images and shares the 1000 classes of ImageNet \cite{deng2009imagenet}. 
We follow \cite{jiang2017mentornet} to create a mini version of WebVision, which includes the Google subset images of the top 50 classes. 
We train all models using the WebVision training set and evaluate on the ImageNet validation set. 
We modify $p=4$ and $k=8$ for mini and $0.4$ for full. 
The default architecture is InceptionResNetv2, the same as compared methods. We also test a smaller ResNet-50.
To create the probe dataset, we split 10 images per class from the ImageNet training data. 
We only observe slight (<0.5\%) gain when we train InceptionResNetv2/ResNet-50 by adding the probe data in training data. As shown in Table \ref{tab:web}, our method significantly outperforms compared methods.

Clothing1M \cite{xiao2015learning} and Food101N \cite{lee2017cleannet} are another two large-scale datasets with real-world noisy labels. We follow their specific settings and train our method to compare with previous methods. Each dataset contains a human verified train subset, which is used as our probe data. 
%Some previous methods use ImageNet pre-trained initialization \cite{han2019deep,guo2018curriculumnet}. 
We use ResNet50 with random initialization. Image size is 224x224. The comparison result of Food101N are shown in Table \ref{tab:food101n}. Our method achieves 77.21\% on the Clothing1M dataset.

\subsection{Comparison to semi-supervised learning}
\label{sec:compare_semi}
We compare our method to one of the advanced semi-supervised learning methods, MixMatch \cite{berthelot2019mixmatch}, and verify how much useful information our method can distill from mislabeled data. 
Figure \ref{fig:overview} shows the comparisons and Table \ref{tab:cifar_semi_experiment} reports the detailed results. 
Given the same trusted set (probe data), our method largely improves the semi-supervised accuracy given 80\% label noise ratio on CIFAR100.
Additionally, the proposed technique (i.e., KL-loss in Section \ref{sec:consistlabel}) improves pseudo labeling so it is supposed to be useful for the compared MixMatch. 
As shown in Table \ref{tab:cifar_semi_experiment}, it is interesting to find out that our extension (denoted as MixMath-KL) shows remarkable benefits for semi-supervised learning, for example, it improves accuracy from 34.5\% to 57.6\%.

\begin{table}[t]
    \caption{Ablation study on CIFAR100. $\checkmark/\xmark$ indicates the corresponding component is enabled/disabled. So M-1 is equal to L2R; M-5 (bold) is the full method.  Abbreviations are defined in text. }
     \vspace{-.3cm}
    \centering
    \begin{tabular}{ccccc|cc}
     \toprule
     \multirow{2}{*}{ M-$\#$ }     &   \multicolumn{4}{c}{Component}     &   \multicolumn{2}{c}{Noise ratio} \\  \cmidrule{2-7}
           & $L_{\text{KL}}$          & $L_\beta$         &  PA & $\lambda$     & 0.4 & 0.8 \\ \midrule
     1     &    &             &              &         &   64.43  & 33.52 \\ 
     2     &    &             &     &  \checkmark &   66.14 & 36.04   \\
     3     &    &             & \checkmark    & \checkmark  &   67.82 & 37.01   \\
     4     &              & \checkmark  & \checkmark  & \checkmark   &  78.06  & 61.81 \\
     5     & \checkmark   & \checkmark &  \checkmark  & \checkmark & \textbf{79.96} & \textbf{75.42} \\ \midrule
     6     &   &     \xmark         &   &  &   73.63    & 54.76  \\
     7     &    &  & \xmark   &   &  79.16 &  72.69 \\
     8     &    &   &   & \xmark   &   81.05  & 74.04 \\ \midrule  \midrule 
     9     & \multicolumn{4}{l|}{ w/o mixup 10 / class} &  78.4 & 72.7  \\
     10    & \multicolumn{4}{l|}{ w/o mixup 1 / class}  &   62.5 & 47.1   \\\bottomrule 
     
    \end{tabular}
    \vspace{-.3cm}
     \label{tab:ablation} 
\end{table}

\begin{figure}[t!]
 \centering
 \includegraphics[width=.8\linewidth]{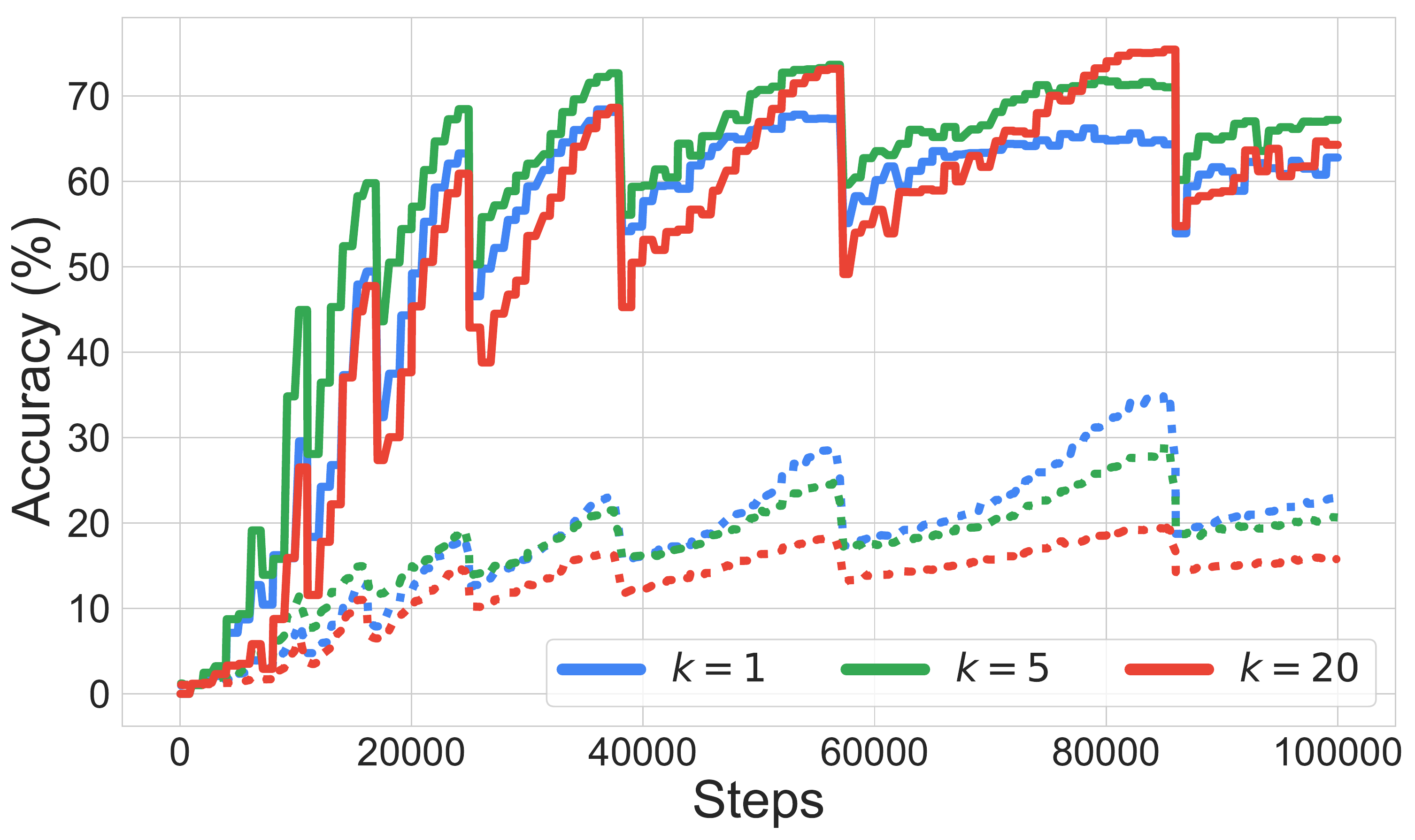}     
 \vspace{-.3cm}
  \caption{Training curves on CIFAR100 with uniform 80\% label noise under different $L_{\text{KL}}$ loss weight $k$ (defined in Algorithm 1). Dotted are solid lines are train and evaluation accuracy curves, respectively.  Since the noise ratio is 80\%, the average training accuracy is expected to be lower than 20\%, otherwise the model starts to overfit. When we use a small $k$, the model becomes to overfit after 70k iterations.}
  \vspace{-.3cm}
  \label{fig:uc}
\end{figure} 

\section{Ablation Studies and Discussions}
\label{sec: discussion}
Here we study the individual objective components and their importance. Table \ref{tab:ablation} summarizes the ablation study results (referred to as M-$\#$) and we discuss them below. 

\textbf{The effects of} $L_{\text{KL}}$:
Based on our empirical observations, $L_{\text{KL}}$ plays an important role in preventing neural networks from overfitting to samples with wrong labels, especially at extreme noise ratios. 
M-4 shows results without $L_{\text{KL}}$. 
Figure \ref{fig:uc} shows the training curves with different coefficient $k$ for $L_{\text{KL}}$. At around 80k iterations, the curve of $\beta=1$ starts to overfit to noisy labels and simultaneously the validation accuracy starts to decrease. $\beta=20$ is much more efficient in overcoming this.

\textbf{The effects of $L_\beta$:}
M-6 shows the result without $L_\beta$. The performance loss is significant at 80\% noise ratio. 
The intermediate step of $L_\beta$ is mixup. It helps the introduction of probe data in supervised training and reduces overfitting (see Section \ref{sec:sup}). M-9 and M-10 study its effect.
If we reduce the probe data size to be 1 sample per class, the accuracy drop becomes significant w/o mixup (the full method with 1 sample per class achieves 75.1\%/62.1\% accuracy with 40\%/80\% noise ratios, as shown in Table \ref{tab:cifar100uniform}). 

\textbf{The effects of data augmentation:} 
The disadvantage of learned PA as used by our method is that it requires learned policies on CIFAR, implying the use of extra labeled data \cite{xie2019uda}. 
We study the contribution of the learned policy to our method with two different experiments.
First, M-3 and M-7 show the results without learned policy augmentation (we only use flip ${\rightarrow}$ random crop ${\rightarrow}$ cutout). The accuracy decrease is minor given 40\% noise and less than 3\% given 80\% noises.
Second, we completely randomize the policies following \cite{cubuk2019randaugment}, we observe that accuracy are almost identical to the original results at all noise ratios. 
The two experiments indicate that our method does not rely on leaned policies and removing them keeps our method effective.

\begin{figure}[t]
    \centering
    \includegraphics[width=.9\linewidth]{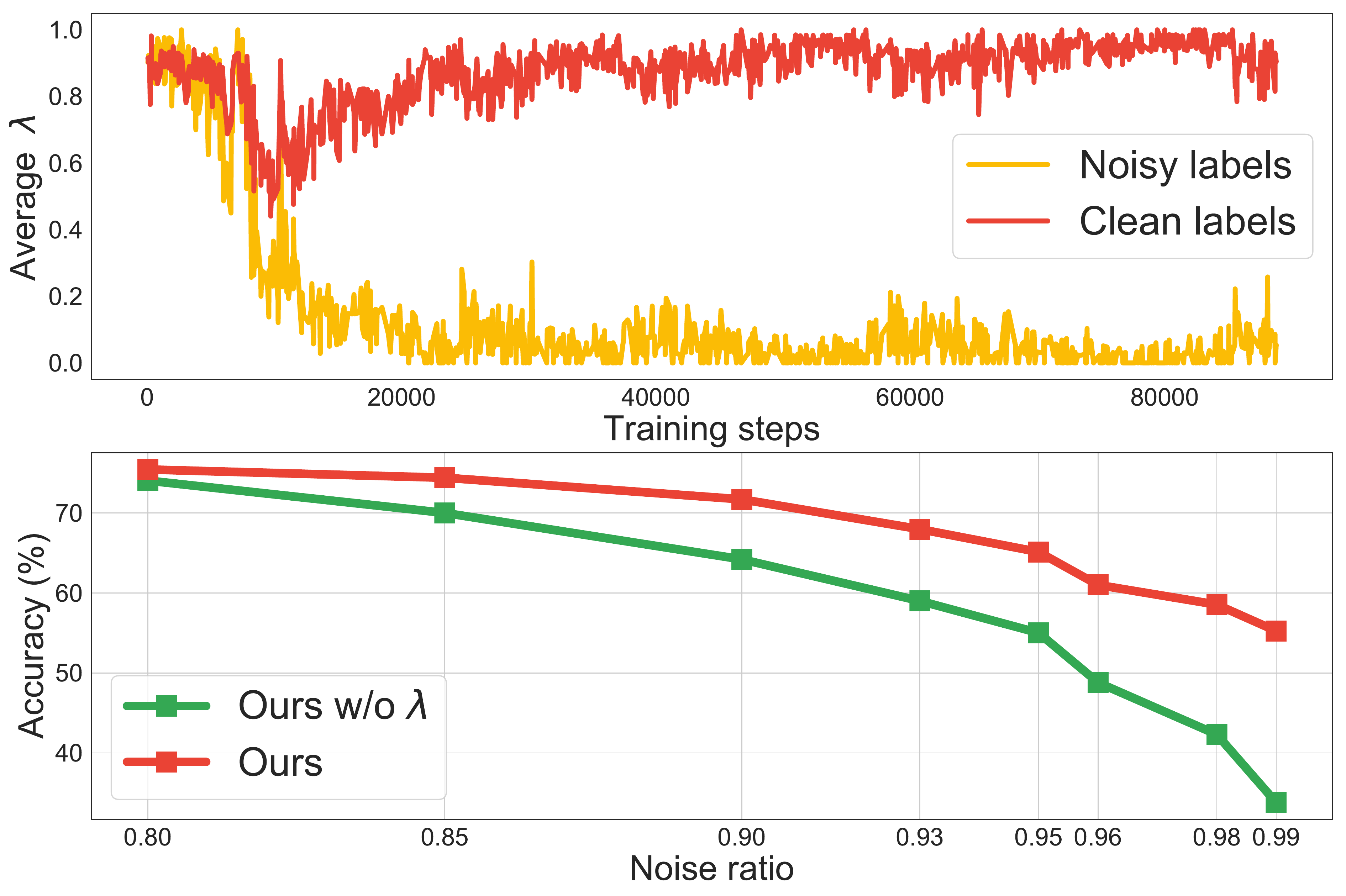}
    \vspace{-.4cm}
    \captionof{figure}{Analysis of $\lambda$. Top: The average $\lambda$ of noisy and clean labels on CIFAR10 with 40\% noise. The average $\lambda$ at 50 epoch converts to ${\sim}0.6$, indicting 40\% mislableled data are detected. Bottom: Accuracy (w/o $\lambda)$ at extreme noise ratios on CIFAR100.}
    \label{fig:lambda}
    \vspace{-.3cm}
\end{figure}

\textbf{The effects of $\lambda$:} 
Our proposed meta re-labeling (Equation \eqref{eq:theta}) is very effective for high noise ratios. We observe comparable performance of models without re-labeling at low noise ratios (e.g. M-5 vs M-8), indicating higher effectiveness of meta re-labeling given higher noise ratios, however, less effectiveness at low noise ratios.
Figure \ref{fig:lambda} (top) shows the average $\lambda$ during the training process (the value of noise labels are obtained by peeping ground truth). 
It learns to reduce $\lambda$ for mislabeled data in order to promote the use of pseudo labels, and vice versa for clean data. 
Figure \ref{fig:lambda} (bottom) demonstrates the significant advantage of the proposed $\lambda$ at extreme noise ratios.

\section{Conclusion}
% \vspace{-.2cm}
We present a holistic noise-robust training method to address the challenges of severe label noise. 
Our approach leverages a small trusted set to estimate the exemplar weights and labels (namely Data Coefficients) and train models in a supervised manner that is highly invulnerable to label noise. 
Comprehensive experiments are conducted on datasets with various types of label corruptions. %Our method sets new state of the art on all tested benchmarks.

Learning from noisy labels is a highly desirable capability. This paper suggests two takeaways. First, small trusted set is not costly but highly valuable to acquire. Designing noise-robust methods that leverage them can have much higher potential to improve performance. 
To the best of our knowledge, this paper is the first to demonstrate superior robustness against noise regimes as high as over 90\%.

\section*{Acknowledgments}
We would like to thank Liangliang Cao, Kihyuk Sohn, David Berthelot, Qizhe Xie, and Chen Xing for their valuable discussions.

{\small
\bibliographystyle{ieee_fullname}
\bibliography{egbib}
}

\newpage
\newpage
\onecolumn

\appendix
\counterwithin{figure}{section}
\renewcommand{\thefigure}{A\arabic{figure}}
\setcounter{table}{0}
\renewcommand{\thetable}{A\arabic{table}}

\section{Proof of small $\nabla\lambda$}
Here we demonstrate that the derivative of $\lambda_t$, $\frac{\partial } {\partial \lambda_{t}} \mathbb{E} \big[L_p|_{\lambda = \lambda_0, \mathbf{\omega} = \mathbf{\omega}_0} \big]$, inside the sign function of Equation 6 will become very small when pseudo labels are close to corresponding original labels. $L^p$ is loss on the probe data $D_p$ with $M$ samples.

\begin{align*}
& \frac{1}{M} \sum_{i=1}^{M} \frac{\partial }{\partial \lambda_{t,i}} L_p(y_i, \Phi(x_i; \Theta))|_{\mathbf{\omega}_{t,i}=\mathbf{\omega}_{0,i}, \lambda_{t,i}=\lambda_{0,i}} \\
& = \frac{1}{M} \sum_{i=1}^{M} \frac{\partial L_p(y_i, \Phi(x_i; \Theta))}{\partial \Theta}|_{\Theta=\Theta_t}^T \frac{\partial \Theta_{t+1 (\lambda_{t,i})}}{\partial \lambda_{t,i}} |_{ \mathbf{\omega}_{t,i}=\mathbf{\omega}_{0,i}, \lambda_{t,i}=\lambda_{0,i}} \\
& \propto \sum_{i=1}^{M} \frac{\partial L_p(y_i, \Phi(x; \Theta))}{\partial \Theta}|_{\Theta=\Theta_t}^T \frac{\partial 
(\mathbf{\omega}_{t,i} \cdot L(y_i, \Phi(x_i; \Theta)) - \mathbf{\omega}_{t,i} \cdot L(g(\Phi(x_i; \Theta), \Phi(x_i; \Theta))
}{\partial \Theta}|_{\Theta=\Theta_t, \mathbf{\omega}_{t,i}=\mathbf{\omega}_{0,i}} \\
& \propto \sum_{i=1}^{M} \frac{\partial L_p(y_i, \Phi(x_i; \Theta))}{\partial \Theta}|_{\Theta=\Theta_t}^T \frac{\partial 
(L(y_i, \Phi(x_i; \Theta)) - L(g(\Phi(x_i; \Theta), \Phi(x_i; \Theta))
}{\partial \Theta}|_{\Theta=\Theta_t} 
\end{align*}

If $y_i$ and $\Phi(x_i; \Theta)$ are close to each other around $\Theta_t$, the derivative $\frac{\partial }{\partial \lambda_{t}} \mathbb{E} \big[L_p\big|_{\lambda = \lambda_0, \mathbf{\omega} = \mathbf{\omega}_0}]$ would be close to 0. Thus, for a converged model with low training error, the norm of gradient on $\lambda$ would be close to zero. The mathematical view motivate our design of Equation 6 instead of $\lambda \approx \lambda_0 - \nabla_\lambda L^p|_{\lambda = \lambda_0}$.

%As can be seen, $\partial (L(y_i, \Phi(x_i; \Theta)) - L(g(\Phi(x_i; \Theta))$ is proportional to $\frac{\partial }{\partial \lambda_{i,t}} \mathbb{E} \big[L_p\big|_{\lambda = \lambda_0, \mathbf{\omega} = \mathbf{\omega}_0}]$. If it is close to 0, the derivative is close to 0.

% $\mathbf{\omega}_{t}^{*}\approx\mathbf{\omega}_{t-1}^{*}-\nabla_{\omega}L^{p}$

\section{Extra training details}
The hyperparameters $p$ and $k$ vary for different datasets, thought we use the identical parameters for all CIFAR results. 
For WebVision mini dataset, we use $p=4, k=40$. For WebVision full dataset, we use a smaller $k=4$ works well. On Food-101N, we set $p=1, k=14$. For Clothing1M, we set $p=1, k=3.5$. Large $k$ on Clothing1M and Food-101N will encourage the model focus on the $L_{\text{KL}}$ too much and yield the convergence issue.
Algorithm 1 step 4 uses a weight threshold $T$ to divide the training batch to possibly clean set and possibly mislabeled set. 
In our experiments, we find setting $T$ to be highest is optimal in terms of training stability, i.e. all data is treated as possibly mislabeled, because it makes the batch size fixed to compute other losses that use data with pseudo labels.

The CIFAR experiments are conducted on a single V100 GPU and all others are conducted on Google TPU with 32 cores.

\section{Training time}
The baseline Learning-to-Reweight (L2R) has theoretically $\sim$3x training time of vanilla training, including a forward pass on the training data, a forward pass on the probe data, and a backward-on-backward pass for computing weights (which takes the same time as a forward pass). 
In analogy to L2R analysis, our method has one more feedforward pass for the augmented input (Section 3) and small loss computation overhead in Eq (3),(4),(8)\&(9). So the total training time is $\sim$4x of vanilla training. Without implementation optimization, we observe $\sim$2x memory and yields ${\sim}3{-}7$x GPU hours of vanilla training across different datasets/architectures.

\end{document}